\documentclass{llncs}
\usepackage{mathptmx}       % selects Times Roman as basic font
\usepackage{helvet}         % selects Helvetica as sans-serif font
\usepackage{courier}        % selects Courier as typewriter font
\usepackage{type1cm}        % activate if the above 3 fonts are not available on your system
\usepackage{makeidx}         % allows index generation
\usepackage{graphicx}        % standard LaTeX graphics tool when including figure files
\usepackage{multicol}        % used for the two-column index
\usepackage[bottom]{footmisc}% places footnotes at page bottom
\usepackage{subfigure}
\usepackage{amsfonts}
\usepackage[cmex10]{amsmath}

\begin{document}

\title{Towards quantitative methods to assess network generative models}
\titlerunning{Quantitative assessment}  % abbreviated title (for running head)
\author{Vahid Mostofi\inst{1} \and Sadegh Aliakbary\inst{2}}
%
%\authorrunning{Ivar Ekeland et al.} % abbreviated author list (for running head)
%	Craig Chambers, Kim B. Bruce, and Elisa Bertino}
%
%\institute{Princeton University, Princeton NJ 08544, USA,\\
%	\email{I.Ekeland@princeton.edu},\\ WWW home page:
%	\texttt{http://users/\homedir iekeland/web/welcome.html}
%	\and
%	Universit\'{e} de Paris-Sud,
%	Laboratoire d'Analyse Num\'{e}rique, B\^{a}timent 425,\\
%	F-91405 Orsay Cedex, France}

\institute{	
	\email{v.mirzaebrahim@mail.sbu.ac.ir},\\ 
	\and
	\email{s\underline{ }aliakbary@sbu.ac.ir},\\
		Faculty of Computer Science and Engineering, Shahid Beheshti University G.C, Tehran, Iran\\
	}

\maketitle    

\section{Introduction}

There is a long-standing history with graph generative models \cite{NetGAN}. Modeling physical and social interactions, discovering new molecular and chemical structures, and constructing knowledge graphs are some example applications of generative models \cite{GraphRNN}. Traditionally, many human-designed generative models (e.g. Barab\'{a}si-Albert model \cite{BA}) are developed to model a particular kind of graphs (e.g. scale-free networks) \cite{GraphRNN}. But a recent and modern approach of network generation methods exploit automatic construction of generative models based on deep learning techniques such as Generative Adversarial Networks (GAN) \cite{GAN} and Variational Autoencoders (VAE) \cite{VAE}. GraphRNN \cite{GraphRNN} and NetGAN \cite{NetGAN} are some recent efforts in employing deep learning for network generation. Since such generators (e.g., GraphRNN and NetGAN) use different techniques of deep learning and result in different generative characteristics, it is necessary to quantitatively evaluate them in order to effectively compare their results. Generally, it is a challenging task to assess the quality of generative models \cite{noteOnEval}. While defining a quantitative assessment for images and text generative models has been a hard task \cite{noteOnEval}, it is more challenging for graph generative models. It is common for scholars to evaluate graph generative models using qualitative techniques, but qualitative evaluation can be very difficult in case of graphs, specially large graphs. It is not easy for an individual to judge a graph unless the graphs are fairly simple and planar \cite{GraphVAE}.

%Each of these methods use their own evaluation technique and mostly just compare basic statistical features of graphs (features like degree distribution, clustering coefficient). Hence evaluating the generated graphs becomes vital as we have all these different ways to do the same task.

In order to design a quantitative approach to evaluate the generated graphs, we studied different approaches to the same problem in other fields such as image and text generation. A common and established approach in evaluating generative models is based on utilizing classifiers \cite{pnGAN}. In this paper, we propose a general quantitative method for assessing graph generative models. The proposed method is used to evaluate and compare modern models which are based on deep learning, but it is not only limited to deep learning techniques. 

\section{Proposed Method}

There are existing efforts in the literature for employing classifiers in order to evaluate a generative model \cite{evaluator_1}. Generative models are designed to synthesize artificial data which are similar to real data. If the generated samples are realistic (similar to real samples), it would be hard for a classifier to distinguish the generated samples from the real ones \cite{evaluator_1}. As a result, if an established and accepted classifier fails to effectively distinguish real and synthesized data, the generative model has performed its job well. In other words, the inaccuracy of an accepted classifier for distinguishing real and artificial samples is a witness of the accuracy of the generation method. 

We propose to use graph classifiers in order to quantitatively assess graph generators. In this regard, we employ Deep Graph Kernels (DGK) \cite{DGK} as an accepted and established classifier in the filed of complex networks, but other accepted graph classifiers could replace DGK if necessary. If the generated graphs are similar to real ones, the classifier would fail to distinguish them and consequently, the accuracy of the classifier would tend to 0.5. In other words, the closer the classifiers accuracy is to 0.5 (i.e., 50 percent precision), the better the generative model has done its task. As a result, we consider the distance of the classifier accuracy from the 0.5 value as a quantitative value for scoring the generative model. If the accuracy of the classifier $c$ is equal to $acc_c$ for distinguishing graphs generated by the generative model $m$ from real graphs, we define the error of the generative model equal to $error_m = |acc_c - 0.5|$.

%For graphs, it requires a comparison between two sets of graphs (the generated graphs and the test set) \cite{GraphRNN}. 

%Existence of many different graph classifiers like \cite{DGK} \cite{Graph2DCNN}, makes using classifiers for evaluating generated graphs a suitable choice. So 
%
\section{Experiments and Results}
As an experiment, we compared NetGAN with different variations of GraphRNN. Deep Graph Kernels (DGK) \cite{DGK} has been used as classifier in all experiments. We configured DGK with MLE kernel, graphlets features and left the rest of the configurations as default. GraphRNN can use different generative models including VAE, RNN and MLP, and we used RNN and MLP in this experiment.

We conduct our experiments on two types of graphs: 1- Caveman graphs \cite{caveman_graphs} 2- scale-free graphs generated by Barab\'{a}si-Albert model \cite{BA}. For each graph type, we consider several graph samples as the ``real graphs'', and then we synthesize similar graphs (with those real graphs as the target networks) using the generative models, and we label the generated graphs as ``fake'' samples. Finally, we feed the ``fake'' and the ``real'' graphs to the DGK classifier. The classifier goal is to distinguish ``real'' and ``fake'' graphs. Therefore, the generative model that results in less real/fake classification accuracy is preferred, because it has been able to fool the classifier, and the classifier has failed to distinguish real and fake graphs.

\begin{table}[htp]
\begin{center}
\begin{tabular}{lccc}
\hline
Garph Type 			& 	GraphRNN:RNN 	&	 GraphRNN:MLP 	& 		NetGAN  	\\ \hline
Barabasi-Albert 	&		71.0\% 	 			&  	78.4\% 				& 		57.0\% 		\\ \hline
Caveman 				& 		88.7\% 				& 		99.6\% 				& 		74.0\% 		\\ \hline
\end{tabular}
\end{center}
\caption{Performance measures of GraphRNN and NetGAN using DGK as classifier,  lower is better}
\label{table:results}
\end{table}

Table \ref{table:results}, shows the accuracy of the DGK classifier in different scenarios. As the results show, NetGAN is better than GraphRNN in generating scale-free (Barab\'{a}si-Albert) and Caveman graphs because the classifier accuracy is more close to 0.5 in both cases. In other words the classifier has failed to to identify ``real'' graphs (target networks) from the ``fake'' graphs (synthesized networks) generated by NetGAN. Additionally, it seems that generating Caveman graphs is a harder task than synthesizing scale-free graphs. Since the classifier shows more precision in the case of the Caveman graphs.

\section{Conclusion}

Generating graphs using deep learning approaches is a new and growing field. Therefore, the need to quantitative evaluation metrics for graph generative models is overwhelming. We proposed to utilize graph classifiers in order to evaluate graph models. Our preliminary experiments show that NetGAN performs better than GraphRNN variations for two considered network types. As the next steps of this research, we will use different generators and classifiers in future. Additionally, real graph datasets will be included in our experiments. 

\bibliographystyle{plain}
\bibliography{assessgenerativemodels}

\end{document}